# Force interaction, modeling and soft tissue deformation during reciprocating insertion of multi-part probe

Tassanai Parittotokkaporn[1*], Matthew Oldfield[1,2], Luca Frasson[1], Ferdinando Rodriguez y Baena[1]

*Abstract*—The bio-inspired engineering of ovipositing wasps, which employ a reciprocating motion for soft tissue insertion, offers potential advantages in reducing insertion force and minimizing tissue damage. However, the underlying mechanisms of tissue interaction and sparing are not fully understood. In this study, we aim to investigate a multi-part probe designed to mimic the reciprocating motion of ovipositors. A reciprocal insertion model was developed to study the interaction between the probe and soft tissue, and experimental testing was conducted using a force sensor and laser optical technique to gain insights into interacting forces and tissue deformation. The results reveal that during the cutting phase of reciprocal motion, the peak force and average displacement of the soft substrate were approximately 19% and 20% lower, respectively, compared to direct insertion at an overall probe velocity of 1 mm/s. This study presents a novel approach combining mechanical modeling and experimental analysis to explore the force mechanics of the reciprocating insertion method, providing a better understanding of the interaction between the probe and soft tissue.

## I. Introduction

Natural inspiration for drilling tools can be found in certain biological organisms, such as the parasitoid wasp, which can penetrate a host larva without damaging the host itself [1–3]. However, little is known about how this mechanism prevents tissue damage while preserving the host. The unique reciprocating mechanism and insertion strategies of the ovipositor have been reported experimentally [4–7] as well as in space and biomedical applications [8,9]. Nevertheless, a deeper understanding of soft tissue responses is necessary to fully comprehend the reciprocating mechanism beyond previous studies [10,11]. Basic knowledge of the reciprocal motion of probe insertion is essential for this purpose. Fig. 1 illustrates the concept of forward motion through a reciprocating mechanism of two-segment probe. The reciprocal motion penetrates tissue by alternately actuating two segments with minimal axial force from its base [12].

Tassanai Parittotokkaporn: tassparitt@gmail.com
[1]*Mechatronics in Medicine Lab, Department of Mechanical Engineering, Imperial College London, London SW7 2AZ, UK*
[2]*Department of Mechanical Engineering Sciences, Faculty of Engineering and Physical Sciences, University of Surrey, Guildford GU2 7XH, UK*

The reciprocal motion of the two segments drives the ovipositor's forward motion, as one segment is pushed deeper into the tissue, stabilized by the tension generated in the other segment. Equation (1) describes this behavior:

$$F_e < F_c + F_i < F_d \qquad (1)$$

Where $F_d$ is the driving force used to push the inward segment of the probe. $F_i$ is the inserting friction force caused by sliding friction against the soft tissue [13]. $F_c$ is the cutting force required at the tip of the segment to cut and displace the tissue [14,15], and $F_e$ is the extracting friction force caused by the segment surface grasping the tissue. Therefore, the stationary segment acts as a gripper, anchoring the soft tissue to allow the moving segment to penetrate.

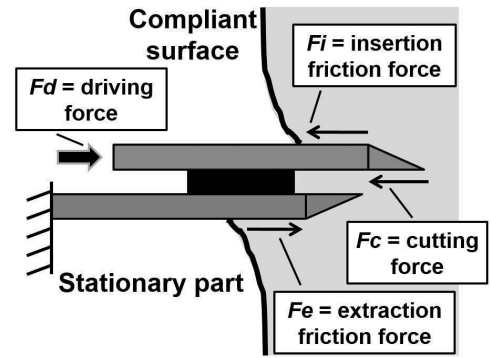

Fig. 1. Reciprocal motion of two segments. The lower segment is kept stationary, functioning as an anchor to the compliant media due to contact friction. This segment serves as a guide for the upper segment that is pushed inward into the tissue. Full arrows represent a push force and thin arrows interacting forces. The extraction forces ($F_e$) keep the stationary part anchored in the substrate and counteract the insertion friction and cutting forces ($F_i + F_c$) of the pushing segment. Inner friction is considered negligible.

To achieve reciprocal penetration, we hypothesize that the extraction friction forces of the stationary segments must be greater than the cutting force plus the insertion friction force of the pushing segment, as shown in Fig. 1. Otherwise, the soft tissue would be deformed and pulled away from the gripping segments. By alternately pushing and pulling each segment, the probe can be inserted into the substrate while avoiding excessive net push forces and axial loads that could damage the tissue [15]. In general, most ovipositors consist of more than two segments. Therefore, a multi-part probe model was developed and tested with soft tissue positioned on a wheel bearing against a force sensor, as shown in the diagram in Fig. 2.

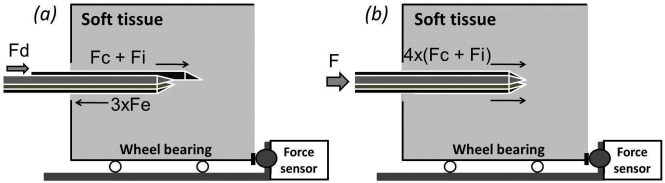

Fig. 2. Schematic of probe insertion by (*a*) reciprocal motion and (*b*) direct pushing into a soft tissue sample placed on a wheel bearing against a force sensor.

The force ($F_d$) applied to push the reciprocal segment into the tissue is subtracted from the extraction forces, resulting in the counteracting force detected by the force sensor. The internal friction of the probe interlocking system and wheel bearing is assumed to be negligible. The pushing force ($F_d$) can be reduced by 1) decreasing the cutting force ($F_c$), including the insertion friction force ($F_i$), by using a sharper tip and smaller size for the probe segment, respectively, and 2) increasing the extraction friction force ($F_e$) by increasing the contact surface area of the stationary segments. Therefore, the multi-part probe was designed with four segments, as shown schematically in Fig. 2(*a*). Hence, $3F_e$ is the force exerted by the probe surface gripping the tissue with the three stationary segments.

$$F_c + F_i < F_d < 3F_e \quad (2)$$

In this study, we aimed to investigate probe-tissue interactions in terms of the magnitude of the force reaction and tissue deformation, comparing reciprocal motion to direct insertion of the four-part probe. In direct pushing, a force of $F = 4\times(F_c + F_i)$ is required to push all segments of the probe, as illustrated in Fig. 2(*b*). As a result, the larger force required to penetrate the whole probe directly may affect tissue deformation. This raises the research question of how the force interaction correlates with tissue deformation, compared between direct insertion and reciprocal penetration of the four-part probe.

Although many studies have reported that bio-inspired needle insertion can lower the insertion force [16] as measured from the back of the needle, none of them have investigated the interacting forces directly measured from the soft tissue in correlation with the tissue deformation. In this study, we first developed a mechanical model to demonstrate reduced force interaction due to reciprocating insertion of a four-part probe. Subsequently, we conducted an experiment to investigate the mechanical characteristics of force interactions, comparing reciprocal and direct probe penetration, using tissue-mimicking gel specimens. Additionally, we explored tissue displacements during needle insertion through a laser-based image correlation technique [10], and analyzed strains near the needle interface to study interactions between a needle and surrounding soft material in both conventional direct insertion and reciprocal penetration [11].

II. Materials and Methods

*A. Model analysis and simulation*

To analyze the performance of the system force interaction, we simulated the reciprocal and direct needle insertion system using Simscape (Simulink 2020b, MathWorks), with a translational friction and spring-damper model as shown in Fig. 3. We used the Modified Kelvin model to represent needle interaction with viscoelastic material [17]. The interaction forces between the needle and the soft tissue, such as the friction force along the needle shaft and the cutting force at the needle tip, were modeled in the translational friction block [18]. Each block represents friction in the contact between a moving single-part probe. The friction force was simulated as a function of relative velocity and assumed to be the sum of Stribeck, Coulomb, and viscous components [19].

*1. Reciprocal motion*

Each of the four-segment probes is reciprocally pushed inward at a constant velocity, exerting a pushing force against the three stationary segments, creating static friction between the probe and the tissue (Fig. 3(*a*)). This static friction is represented as a stuck box and results in less deformation of the tissue surrounding the probe during the compression phase. As each probe is sequentially pushed deeper, the static friction transitions into dynamic friction when the probe slides at a different velocity relative to the adjacent probe segment and surrounding tissues, represented by the slip boxes (Fig. 3(*b*)). The tip of the probe segment cuts through the soft tissue, causing irreversible displacement and disruption of the tissue during the cutting phase.

*2. Direct insertion*

During the compression phase (pre-sliding), the combined four-segment probe is pushed inward at a constant velocity, and the static friction force between the probe and the tissue matches the pushing force. This static friction is represented as a stuck box in the transitional friction model, causing the tissue surrounding the probe to deform with probe displacement following the viscoelastic model (Fig. 3(*c*)). In the cutting phase (sliding), the combined probe is sequentially pushed deeper, transitioning the static friction into dynamic friction as the probe slides at a different velocity relative to the surrounding tissues. This transition is depicted by the slip box in the transitional friction model shown in Fig. 3(*d*). Simultaneously, the tip of the probe cuts through the soft tissue, resulting in irrecoverable displacement and tissue disruption due to tissue traversal.

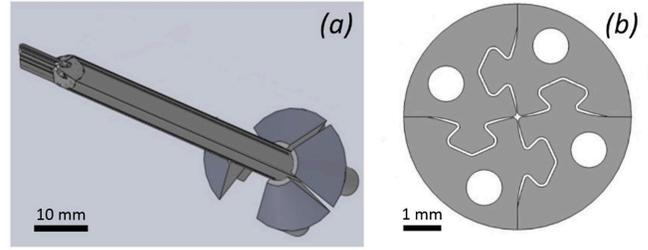

Fig. 4. (*a*) CAD model of a four-part probe with a smooth surface and (*b*) a cross-section of the four-part probe with a diameter of 6.0 mm and a clearance gap of 0.15 mm.

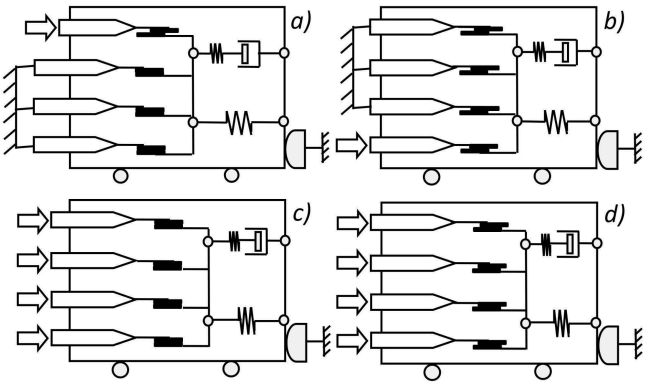

Fig. 3. Sliding friction and viscoelastic model of a four-part probe employing (*a and b*) reciprocal motion and (*c and d*) direct probe insertion.

### B. Probe fabrication

A set of four-segment rigid probes was rapid prototyped from semi-transparent material (FullCure720) with the following material properties: tensile strength of 60.3 MPa, modulus of elasticity of 2,870 MPa, and elongation at break of 20%. After assembly, the probe measured 6.0 mm in diameter and 130 mm in length. Along its length, each segment was equipped with an interlocking rail with a 0.15 mm clearance gap (Fig. 4). The rigid probe was designed with a conical tip, cylindrical shape, and smooth surface. Each segment was moved either simultaneously or reciprocally by a robotic actuator connected to a transmission rod, providing a linear displacement accuracy of 0.01 mm.

### C. Sample preparation

Gelatin was selected as a soft tissue phantom due to its mechanical similarity to human brain tissue [20]. The transparency of gelatin also provides specific advantages for studying tissue deformation. To prepare the gelatin sample, 6 grams of gelatin powder was mixed with 100 cc of hot water (100ºC) and stirred until the solution became clear. Then, 0.10 grams by weight of aluminum oxide particles with an average diameter of 5 μm were added to the liquid gelatin and stirred to obtain a homogeneous solution. The mixture was poured into a transparent polyacrylic box measuring 28.5 x 61.8 x 86.2 mm, with an aperture (12 x 30 mm) designed into the front wall to allow the probe to pass through the sample. The opposite wall was left open to fill the gelatin mixture and to allow the laser beam to project onto the gelatin sample. The transparent box, with 1 mm-thick walls, provided optical access for recording the displacement of microparticles using a high-resolution camera (14-megapixel CCD camera). The mixture was allowed to solidify at room temperature (25.0ºC) for 1 hour, and then kept inside a 4 ºC refrigerator for 12 hours. Afterward, the sample was brought back to room temperature for 2 hours before conducting the experiment.

### D. Actuator and force sensor set-up

Four DC motors with embedded encoders (Maxon Amax22, Model # 300679, 4.4 gear ratio, Maxon Motors Inc.) were used to move each segment of the probe via shape-memory alloy (SMA) rods with a diameter of 1.26 mm [21]. The motors were controlled by a motor-driven controller (CompactRio, LabView v8.6, NI Inc.). A probe trocar was attached to the frame of the actuator to guide the probe through the gelatin sample. A wheel bearing assembly was placed between the box and the frame to minimize friction. A force sensor (Honeywell FSS-SMT Series 1 axis Force Sensor) with a 0-14.7 N operating force range, 12.2 mV/N sensitivity, 1% error, and 1,000 Hz sampling rate was attached to the opposite side of the frame [22]. The interacting force was measured horizontally in the direction of the probe insertion towards the gelatin sample. The force was measured directly from the force sensor as it touched the acrylic box while the probe was pushed into the gelatin sample, as shown in Fig. 6.

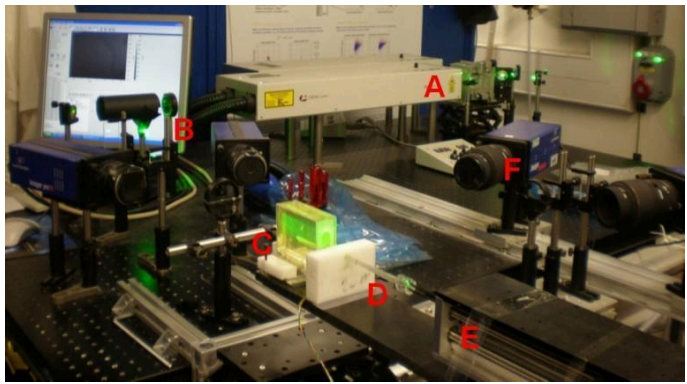

Fig. 5. The experimental setup: (*a*) Nd-YAG laser source, (*b*) optical lens and mirrors, (*c*) tissue sample inside a clear plastic box, (*d*) a four-part probe passing through a trocar, (*e*) a reciprocal actuator, and (*f*) a high-resolution CCD camera.

### E. Particle image velocimetry (PIV)

The laser experiment involved a Nd-YAG laser emitting 10 ns pulses at a wavelength of 532 nm, and a CCD camera (LaVision Imager Pro X 14M) was used for image acquisition [23]. The optical setup is schematically shown in Fig. 5, illustrating the positioning of the sample, needle, and measurement plane. The laser beam was shaped into a light sheet with a thickness of 100 µm and a height of 30 mm. The open side of the sample box was oriented towards the laser beam to eliminate reflections from the acrylic walls. The actual field-of-view captured by the camera was 27.58 x 39.42 mm, with an optical resolution of 27.4 µm/pixel. The tip of the needle was inserted toward the middle of the box and entered the field-of-view at a height of 13 mm above the bottom of the image boundary. The distance between the probe aperture and the image boundary was 18.5 mm, as shown in Fig. 6.

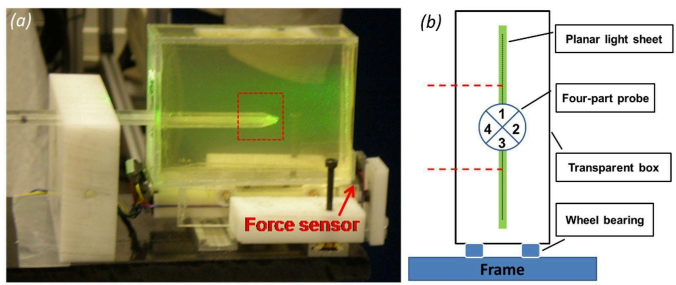

Fig. 6. (*a*) Laser light sheet was projected through the gelatin sample while the four-part probe was in motion. The scattering of green light from microparticles was visualized and focused on the target area measuring 27.58 x 39.42 mm (dashed square). (*b*) An axial view showed the orientation of the four-part probe.

### F. Testing protocol

The testing protocol was divided into 1) direct insertion, the entire segment of the four-part probe was directly inserted into the samples at the speed of 1 mm/s, with a probe displacement of approximately 70 mm from the sample surface and 2) reciprocating motion, each segment of the four-part probe was displaced reciprocally towards the sample. Each part was driven at a speed of 4 mm/s and a displacement of 5 mm in each direction, without any pause between the movements. One cycle of reciprocal motion provided 5 mm of total probe displacement within 5 seconds, which was the same speed as direct insertion (1 mm/s). Therefore, reaching a probe displacement of 70 mm required 14 cycles of reciprocal motion. The testing protocol was performed with 6 insertions for each method using new gelatin samples. Additionally, a slower speed of reciprocal motion (1 mm/s) was added to evaluate force measurements at different speeds.

## III. Results

### A. Force simulation results

The simulation model was developed to analyze the system force interacting performance of probe penetration into soft tissue [24], comparing the effects of direct insertion and reciprocal motion. The purpose of the model was to support our hypothesis rather than to perfectly fit the experimental data. The results depicted in Fig. 7(*a*) demonstrate that during the penetration of soft tissue with a similar displacement of 70 mm, the reciprocal motion of the probe transmitted less force compared to direct insertion. Specifically, the tissue absorbed 11.43% less peak force during reciprocal motion than during direct pushing (Fig. 7(*a*)).

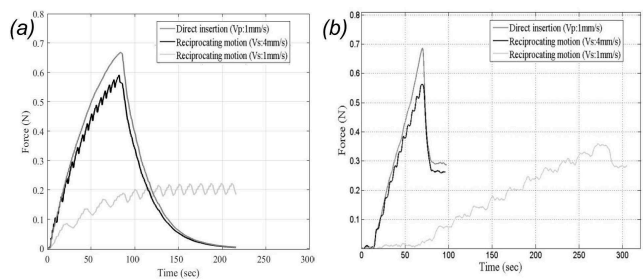

Fig. 7. (*a*) Force profiles obtained from Simscape simulation of a sliding friction and viscoelastic model of a four-part probe. (*b*) Force profiles compared between direct insertion at the speed

(Vp) of 1 mm/s and reciprocating motion at the speeds (Vs) of 4 and 1 mm/s of the four-part probe.

## B. Force measurement results

Commonly, force measurement during probe insertion was done by measuring the force at the proximal part of the probe. However, it is important to note that forces distributed along the entire length of the probe are influenced by various factors, including cutting force, friction force, and forces related to tissue deformation and fracture [25]. Measuring the forces resulting from the alternating motion of a four-part probe could be challenging. In this study, an indirect measurement approach was adopted, where the reaction force on the gelatin box was simultaneously measured alongside the study of tissue deformation using laser-based image correlation. The force sensor was positioned at the back of the sample box to measure the reaction force, which had the same magnitude as the net force acting on the probe but in the opposite direction [26].

Fig. 7(*b*) depicts the average force profiles during direct insertion and reciprocating motion, both showing an increase in amplitude over time and similar relaxation behavior. However, it was important to note that initially, the force sensor did not detect force when the probe came into contact with the tissue. This lack of initial force detection could be attributed to the elastic deformation of the tissue. Approximately 7 seconds into the insertion, the force profile began to steadily increase until it reached a maximum point when the probe reached its end position. Subsequently, the force relaxed to a certain level. It was worth mentioning that this measurement captured the overall force acting on the entire sample, rather than a specific point.

Table 1 summarizes the force measurements, which show a comparison of the average maximum interacting forces transferred to the soft tissue during direct and reciprocating probe insertions at the depth of 70 mm. The results indicate that direct pushing of the probe results in a larger amount of force and energy being transferred to the tissue (0.69 ± 0.04 N and 25.23 ± 0.73 mJ). In contrast, the soft tissue experienced forces and energy of only 0.56 ± 0.08 N and 19.69 ± 0.97 mJ, and 0.37 N and 12.92 mJ during reciprocating motion at the speeds of 4 mm/s and 1 mm/s, respectively. Thus, decreasing the speed of reciprocating motion resulted in a reduction in the interacting force. Additionally, reducing the probe velocity of direct insertion was also expected to decrease the insertion force. However, a probe velocity of 0.25 mm/s for direct insertion was not tested.

TABLE I
Average peak forces and transferred energy during probe-tissue interaction obtained from direct insertion and reciprocating motion of the four-part probe.

| Insertion methods | Average peak force (N±SD) | Work (mJ±SD) |
| --- | --- | --- |
| Direct insertion (Vp: 1mm/s) | 0.69±0.04 | 25.23±0.73 |
| Reciprocal motion (Vs: 4mm/s) | 0.56±0.08 | 19.69±0.97 |
| Reciprocal motion (Vs: 1mm/s) | 0.37 | 12.92 |

## C. Deformation measurement results

Fig. 8 shows an example plot of tissue deformation, where vectors represent the displacement of microparticle clusters as the needle touched the sample at the box aperture and penetrated through the sample for a distance of 70 mm. Each vector represents the displacement of the soft tissue, with the shape of the needle was masked on the displacement profiles. The microparticle-loaded gelatin sample underwent deformation away from the needle in both the radial direction (Y-axis) due to the needle volume and in the axial direction (X-axis) due to the probe trajectory. In Fig. 8 (*b*), large displacements can be seen on the areas surrounding the probe from the vicinity towards the tip, while small displacements occur further away from the probe. Therefore, the displacement vectors were analyzed separately in the X and Y directions and compared between direct insertion and reciprocal motion.

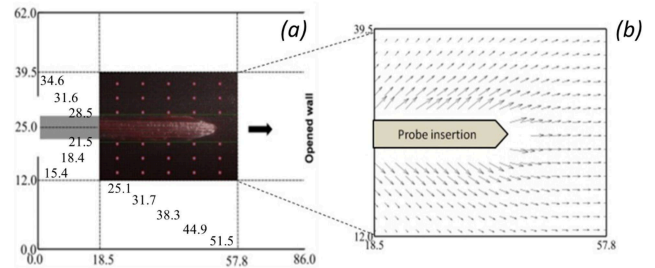

Fig. 8. (*a*) A tracking matrix of microparticle clusters marked in red dots [5 x 6] along the probe shaft. (*b*) Tissue displacement vector were measured as the probe passed through the gelatin sample inside the box. The probe was inserted through the aperture of the box from left to right, with a focused area at the center of the sample box.

### 1. Direct insertion

The X displacement profiles of the five axial locations adjacent to the probe surface were plotted in Fig. 9(*a-b*). These profiles exhibit similar patterns, starting with linear increases in displacement (deformation phase) until reaching a peak, followed by a plateau (cutting phase) where the displacements remain constant until the probe stopped moving. Afterward, the displacements immediately drop to their respective levels (relaxation phase). Notably, at points [5,3] and [5,4], which are further away from the probe tip (Fig. 9(*a*)), there is a large deformation phase but a short cutting phase due to the tissue being deformed towards the opened wall. Conversely, at points [1,3] and [1,4], near the probe entry zone (Fig. 9(*a*)), there is a long cutting phase but a small deformation phase. Additionally, the displacement profiles do not return to their initial positions, with deviations proportional to peak displacement amplitudes. The Y displacement profiles, plotted in Fig. 9(*c-d*), illustrate the radial displacement profiles resulting from the compression of the probe volume against the tissue sample. Fig. 9(*d*) with Y displacement profiles of the five locations along the probe surface, exhibit deformation, compression, and relaxation phases that are distinct from the X displacement profiles in Fig. 9(*b*).

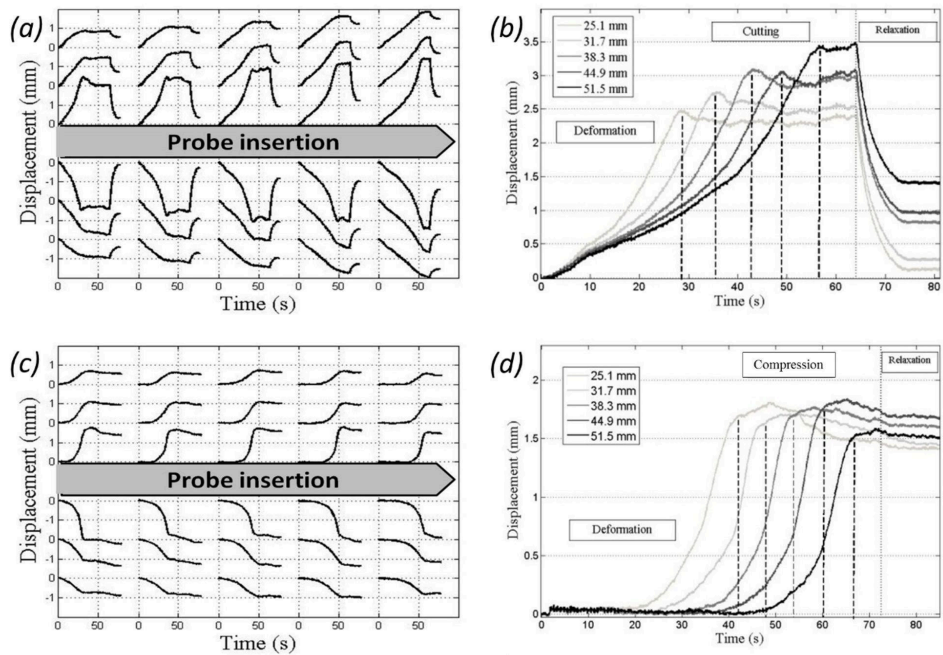

Fig. 9. Amplitude plots of the displacement vector profiles during direct insertion with a probe velocity of 1 mm/s. (*a*) The displacement profiles along the X direction versus time. (*b*) The axial displacements (X) at five locations near the probe surface. (*c*) The displacement profiles along the Y direction versus time. (*d*) The radial displacements (Y) at five locations near the probe surface, with a distance increment of 6.6 mm.

### 2. Reciprocal motion

Fig. 10 shows the displacement profiles of vector displacements in the X direction resulting from the reciprocating motion of the four-part probe. These displacement profiles are nearly similar to the results from direct pushing as seen in Fig. 9, but with reciprocal waves superimposed on the displacement profiles, particularly during the plateau (cutting) phase. Similar to direct insertion, the amplitude of displacement patterns also increases from the location of the probe-entry zone toward the opened wall of the sample box due to tissue elasticity. Again, the five axial locations along the probe surface by 0.5 mm radial distance are plotted in Fig. 10(*b*) which demonstrates consistent patterns, with regions farther away from the probe trajectory exhibiting more deformation phase and less cutting phase compared to areas near the probe entry zone. The superimposed reciprocal wave, characterized by a longer duration and greater prominence, is observed in the position near the probe entry zone.

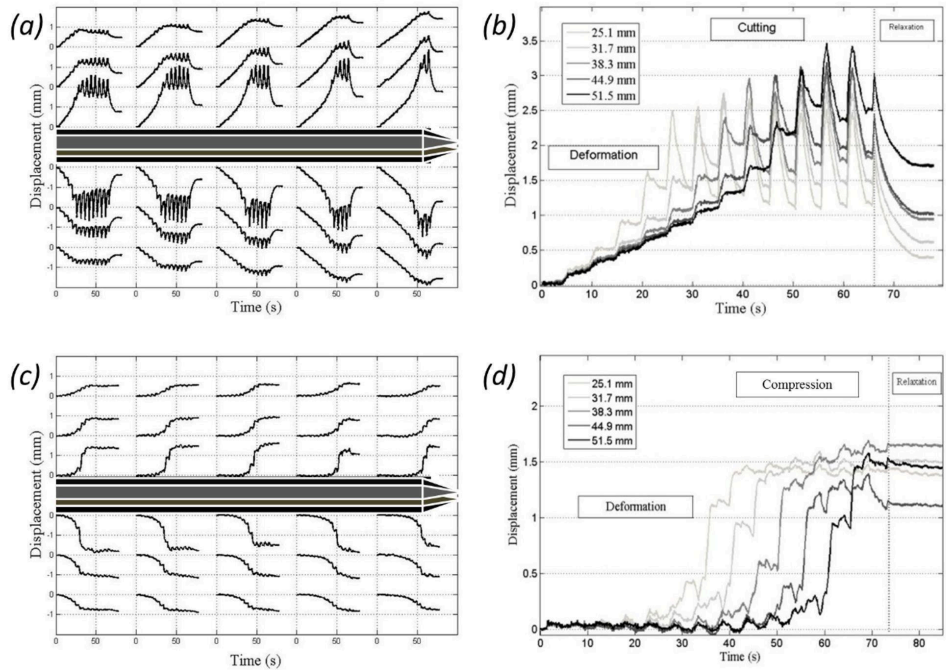

Fig. 10. Amplitude plots of the displacement vector profiles during reciprocal motion with an overall velocity of 1 mm/s. (*a*) Displacement profiles along the X direction versus time. (*b*) Axial displacements (X) at five axial locations along the probe surface. (*c*) Displacement profiles along the Y direction versus time. (*d*) Radial displacements (Y) at five axial locations along the probe surface, with a distance increment of 6.6 mm.

The Y displacement profiles resulting from reciprocating motion are presented in Fig. 10(*c*) and (*d*). As anticipated, the sample experiences compression in the radial (Y) direction as the probe penetrates through it reciprocally. The Y displacements gradually increase at the locations where the probe passes at different time intervals. However, the magnitudes of Y displacements during the compression phase of reciprocating motion (below 1.5 mm) in Fig. 10(*d*) are slightly lower than those during direct pushing (above 1.5 mm) in Fig. 9(*d*). Furthermore, the oscillation patterns resulting from reciprocating motion in the Y direction are less pronounced than in the X direction because the reciprocating trajectory of the probe segments was in the X direction. The oscillating profile pattern on the X displacement is more pronounced during the plateau or cutting phase, rather than the deformation and relaxation phases, as depicted in Fig. 10(*a*). The various phases of the X displacement can be observed in Fig. 11(*a*) between the lower [3,3] and upper [3,4] locations along the probe shaft, based on the alternating tissue displacement.

### D. Force and displacement correlation

An example of X displacement profiles for two locations, [1,3] and [5,3], is shown in Fig. 11(*b*). The profiles obtained from direct insertion and reciprocal motion exhibit the same three phases of deformation, cutting, and relaxation. However, it is noticeable that the average displacement profiles during the cutting phase obtained from reciprocal motion are lower than those from direct insertion. Additionally, the peak displacement of tissue during reciprocal motion still falls within the upper range of the results obtained from direct insertion.

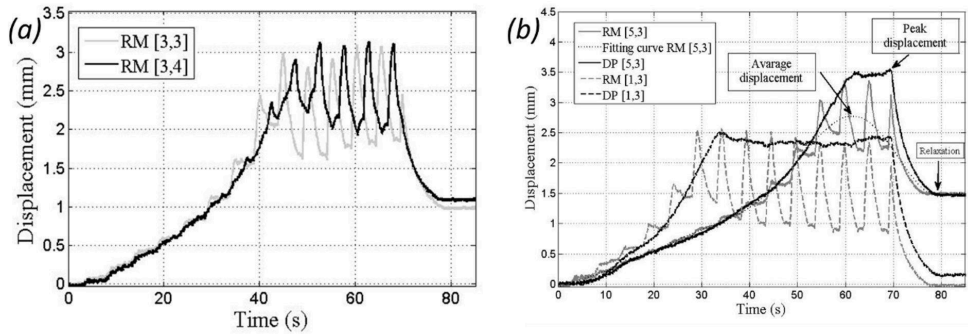

Fig. 11. (*a*) X displacement profiles of two opposite positions across the probe during reciprocating motion with alternating phases of displacements. (*b*) The X displacement profiles of reciprocating motion (RM) and direct push (DP) at the entry point [1,3], the far end point [5,3], and a fitted average curve of a displacement profile (RM [5,3]).

The numerical data were obtained from the same location in the X displacement at [5,3], which is located at the far end and close to the probe surface, resulting in the largest displacement amplitude. However, there is not a significant difference in the peak and relaxation displacements of the soft tissue between direct pushing (Vp: 1 mm/s) and reciprocating motion (Vs: 4 mm/s). Here, Vp refers to the velocity of the probe and Vs refers to the velocity of the segment. However, during the plateau phase of the displacement profiles shown in Fig. 11(*b*), reciprocating motion can result in lower tissue displacement compared to direct insertion (2.92±0.14 mm and 3.63±0.25 mm, respectively). Additionally, reducing the speed of reciprocating motion (Vs: 1 mm/s) leads to a slight decrease in tissue displacement (2.24±0.17 mm), as reported in Table 2.

TABLE II

Numerical data of peak and relaxation X displacements, including the average displacement during the plateau phase at location [5,3], obtained from direct pushing (Vp: 1 mm/s) and reciprocating motion (Vs: 4 and 1 mm/s).

| Insertion methods | Peak displacement (mm±SD) | Relaxation displacement (mm±SD) | Average displacement (mm±SD) |
|---|---|---|---|
| Direct pushing (Vp:1 mm/s) | 3.68±0.28 | 1.78±0.13 | 3.63±0.25 |
| Reciprocating motion (Vs:4 mm/s) | 3.53±0.19 | 1.65±0.09 | 2.92±0.14 |
| Reciprocating motion (Vs:1 mm/s) | 2.96±0.21 | 2.29±0.18 | 2.24±0.17 |

IV. Discussion

In our study, we employed simulation modeling to construct a viscoelastic model (Fig. 3) using an analogous mechanical system in Simscape. To capture the probe insertion behavior, we divided it into three distinct phases: compression, cutting, and relaxation, mirroring our experimental setup. This behavior was meticulously modeled by integrating four translational friction elements for each probe segment, coupled with a viscoelastic model representing the properties of the soft tissue. The deformation at the probe's tip during cutting was simulated using a set of parallel translational friction elements, depicted as sliding boxes. During traversal, as the first probe segment moved, the other three segments held the tissue model in position. This allowed the moving segment to overcome the resistance from static friction in the translational friction elements and effectively penetrate the soft tissue. Eventually, the moving segment reached an endpoint and stopped, compressing the viscoelastic element. This phenomenon resulted in tissue relaxation, which reduced the amount of force required for the subsequent insertion of the probe. Furthermore, this probe-tissue interaction model was able to demonstrate the force profile, which was both explainable with the experimental data.

The core objective of our study was to validate the hypothesis that the reciprocal insertion method mitigates tissue disruption and deformation by minimizing the application of pushing and interacting forces. The reciprocating motion employed by each segment ensured that the cutting force experienced at the segmental tip remained comparatively smaller than the force exerted by the entire probe. This force was effectively counteracted by the gripping forces applied by the other segments. The probe's design, featuring beveled ends, led to an interesting phenomenon whereby the bevels became more pronounced as the probe extended beyond the stationary segment. This tip asymmetry facilitated segment curvature, consistent with prior studies [27,28]. Additionally, the probe's surface topography facilitated tissue gripping while the segment was pulled, consistent with findings in earlier studies [4,5,29]. However, it is worth noting that our study did not delve into the intricacies of the asymmetrical bevel tip resulting from segments offsetting during curved trajectories [30], nor did it explore internal friction forces preventing probe buckling. Furthermore, limitations in the resolution and sensitivity of the force sensor and the image quality of the PIV technique precluded the examination of downsizing the probe, such as employing a 2.5 mm diameter [31].

While the PIV technique proved to be a dependable and non-invasive method for measuring instantaneous soft tissue deformation during probe insertion [10,32], it does possess certain constraints. These include the necessity for intricate calibration, complex algorithms, and a limited camera view. Moreover, the method necessitates transparent samples like gelatin enclosed within a clear acrylic box. In our study, we harnessed the scattering characteristics of laser light from the gelatin sample by introducing microparticles inside the gelatin. The scattered light was then captured and recorded with a CCD camera to analyze soft tissue deformation. Our results affirm the PIV method's precision in measuring displacement vectors, with the magnitude of vector displacement at each region of interest [5 x 6] consistently replicable and closely correlated with directly measured interacting forces from the sample.

We observed substantial displacement regions primarily at the distal area of the probe trajectory and in proximity to the probe shaft. In contrast, pronounced cutting regions were predominantly identified at the proximal area of probe entry. Our analysis demonstrated that the X displacement of the vector field, representing the axial probe trajectory, offered a more suitable parameter for assessing tissue deformation compared to the Y direction, which corresponded to tissue compression due to the penetrating probe. In our comparative analysis, the most significant distinction was evident during the cutting phase, rather than during the deformation and relaxation phases. During the cutting phase of reciprocal motion, the peak force and average displacement of the tissue were approximately 19% and 20% lower, respectively, in comparison to direct insertion. This observation was made while maintaining an overall probe velocity of 1 mm/s. Notably, slower reciprocating speeds further reduced interacting forces and energy transfer to the sample. These findings provide compelling evidence that the reciprocal mechanism of the multi-part probe significantly modifies the energy transferred to the soft tissue, resulting in reduced deformation and, potentially, less tissue damage during insertion.

## V. Conclusions

The concept presented in this study demonstrates the feasibility of mechanical modeling and experimental testing of a multi-part probe inspired by ovipositing wasps. The probe utilizes a reciprocating motion, resulting in minimal interacting force, reduced energy transfer, and minimal tissue deformation during insertion into soft tissue. The results highlight the advantages of the biologically-inspired reciprocating motion method and suggest its potential for the development of a less invasive approach to probe insertion in soft tissue.

## ACKNOWLEDGEMENTS

We would like to acknowledge Dr. Johanes Kerl and Professor Frank Beyrau from the Department of Mechanical Engineering, Imperial College London for their assistance in setting up the pulse image velocimetry.